# ВОЗМОЖНОСТИ ТЕХНОЛОГИЗАЦИИ ФИЛОСОФСКОГО ЗНАНИЯ

**С.Б. Куликов, Томск, Россия**

Цель статьи – анализ вопроса о возможности технологизации философского знания. Под технологизацией мы понимаем организацию познавательной деятельности, которая ориентируется на совокупность приемов, гарантированно приводящих к успешным (т.е. точно соответствующим заданным параметрам) прикладным результатам.

В современной науке актуальность вопроса о возможности технологизации философского знания обусловлена особой проблемной ситуацией, возникающей в свете понимания философии как в целом малоэффективного знания. Например, С. Вайнберг называет это «непостижимая неэффективность философии», отмечая в основном негативное влияние философии, равно как социологии и антропологии на научное мировоззрение и общественное мнение [1, с. 131-149]. Вместе с тем возникает весьма серьезное противоречие. С одной стороны, в число задач философии входит рефлексия, постановка вопросов и раскрытие общих перспектив, но не выработка прикладных решений и потому оказываются возможны негативные оценки эффективности. С другой стороны, очевидно влияние философии на развитие и видоизменение человеческой цивилизации, порождающего положительные эффекты, например, в области образования и политики.

В рамках образования, на различных этапах его развития, ведущую роль сыграли педагогические идеи Платона и, особенно, Аристотеля (в эпоху Средневековья). Как утверждает А. Койре, «труды Аристотеля образуют настоящую энциклопедию человеческого знания. … Не удивительно, что для раннего средневековья, ослепленного и раздавленного этой массой знаний … Аристотель стал представителем истины, вершиной и образцом совершенства человеческой природы, князем тех, кто знает – по выражению Данте, - и особенно тех, кто обучает» [2, с. 55-56]. В свою очередь Ян Коменский в Новое время, опираясь на принцип природосообразности, разработал основы классно-урочной системы, ориентированной на изучение элементов фундаментальных знаний [3, с. 233-259, 263-278].

В отношении политики именно философами были предложены принципы разделения властей и представительной демократии, лежащие в основе современных демократических систем. Например, Дж. Локк обосновал положения: 1) «законодательную и исполнительную власть часто надо разделять»; 2) «у народа верховная власть устранять или заменять законодательный орган, когда народ видит, что законодательная власть действует вопреки оказанному ей доверию». [4, с. 347, 349].

Философия способна порождать конкретные приемы изменения состояний человека и общества, меняя их естественное состояние и придавая черты искусственных объектов. Следовательно, философия может являться основой технологических процессов и приносить пользу. В то же время реализация идей, предложенных философами, растягивается на столетия, полезным же признается обычно то, что может быть применено для какой-либо конкретной цели, т.е. «здесь и сейчас». В связи с этим раскрывается парадокс: философия в общем и целом (т.е. «как таковая») полезна, ибо в конечном итоге находятся способы реализации ее наработок, но в каждый конкретный момент времени она не может быть применена и представляется бесполезной.

Таким образом, необходимость обсуждения вопросов, связанных с возможностями технологизации философского знания, обусловлена задачей поиска особой интерпретации философии, которая позволяла бы учесть парадоксальное единство эффективности и неэффективности данной области знания.

Мы не можем всецело поддержать идею, согласно которой философия едва ли не вся представляет совокупность технологий [5], но как минимум одна из ветвей философии, а именно философия науки способна породить специфическую технологию. Эта технология ориентируется на повышение качества подготовки специалистов. Остановимся на ней более подробно, учитывая, что во многом аналогичные процедуры возможны в отношении развития на базе философии науки ценностных оснований физики и правоведения [6], а также гуманистического потенциала биологии и психологии [7].

Философско-научная технология повышения качества подготовки специалистов включает два компонента:
1. Условия возможности развития методологической культуры.
2. Принципы упорядочивания и усиления методологической культуры [8].

Смысл условий возможности развития методологической культуры раскрывается в двух положениях:
1. Философия науки – это символизация функционирования отличной от неё сферы действительного, в частности, отображенной в логической форме.
2. Символизм философии науки раскрывает возможность понимания логики с позиций многоуровневой системы интерпретации, которая раскрывает представление локальных элементов в виде комплекса априорно-апостериорных метарегулятивов согласования познавательных операций (законов, правил и т.д.) по принципу истинностного мультисигнификатизма (многозначности логических систем) [9; 10].

Данные положения концептуально восполняют результаты, полученные исследователями в области квантовой логики и в обобщенном виде представленные В.Л. Васюковым [11].

Также могут быть выделены особые принципы упорядочивания и усиления методологической культуры:
1. Принцип отраслевой комбинаторики, на основе которого привлекаются различные отрасли знания и выявляется методологический потенциал их взаимодействия.
2. Принцип позитивной парадоксальности, на базе которого философия понимается как неустранимо противоречивое знание, призванное

порождать парадоксы в целях выявления нетривиальных решений традиционных задач [12; 13].

Применение первого принципа возможно, например, в ходе выделения методологических возможностей взаимодействия философии и педагогики в поле действия функционального подхода к словообразованию. Функциональный подход активно разрабатывается в современной филологии, интерпретируясь как «один из вариантов синхронного системного исследования языка» [14, с. 3]. Философский уровень понимания гарантирует присутствие критицизма и рефлексии в границах любой области знания, основания и перспективы которой подвергнуты осмыслению [15, с. 103]. Таким образом, выделение методологических возможностей функционального подхода к словообразованию позволяет раскрыть сущность связи, возникающей между философией, педагогикой и филологией, и предполагает ориентацию на проблематику рефлексии роли языка в научном познании образовательного процесса. Реализация этих возможностей обусловливает поэтапное включение учащихся в процессы получения и производства знаний.

Аналогичные исследования, проводимые в отношении научного языка в рамках филологии [16; 17], а также в отношении технологизации языкового образования [18] не вполне рефлексивны. Вследствие этого затрудняется осмысленность их практического применения, ибо становится мало критичным понимание оснований и перспектив. Междисциплинарный характер исследования, предметом которого выступают условия и предпосылки развития методологической культуры, выгодно отличается тем, что позволяет отрефлексировать роль языка в научном исследовании образовательного процесса, а также раскрыть фундаментальные основания развития данного процесса.

Некоторые специалисты (например, З.И. Резанова вслед за Е.С. Кубряковой, Ю.С. Степановым, М.Н. Янценецкой и др.), подчёркивая целостную иерархическую системность языка, в которой внешнесистемные связи (функции) определяют внутреннее устройство каждого уровня, выделяют следующие признаки функционального подхода:
1. Термин «функция» синонимичен термину «отношение».
2. Характеристика непосредственной функции элемента в пределах смежного уровня (фонемы – в морфеме, морфемы – в слове, слова – в предложении) должна вестись с учётом модификаций этой функции в процессе включения единиц в ярусы языковой системы.
3. Функциональные связи получают завершённость в рамках языка как коммуникативного аппарата в целом [14, с. 3-4].

Таким образом, суть функционального подхода к словообразованию заключается в том, что образование слов понимается производным от сложно организованной системы отношений уровней языка. Каждый уровень имеет собственные функции, относительно независимые от целого. В то же время все они подчинены главной направленности языка на осуществление коммуникации и служат созданию условий понимания сообщений как в целом, так и в частностях.

Соотношение функционального подхода и педагогики выявляет:
1. Различение языка учителя (преподавателя училища или вуза) и языка ученика (студента) по уровню в связи с возрастной дистанцией, спецификами социального положения и др., с одной стороны, а также элементов коммуникативного аппарата языка в филологии – с другой.
2. Иерархическая организация образовательного процесса как формирования учащимися представлений о мире в процессе усвоения элементов сообщения (соответственно букв алфавита, слов, предложений, текстов) через отражение в этих элементах некоторых понятий (в букве – звука, в слове – значения, в предложении – смысла высказывания и т.д.), в ходе восприятия сложносоставных блоков сообщений (например, литературных образов), а также при выстраивании ответной реакции (устных ответов, письменных сочинений, квалификационных и научно-квалификационных работ и др.).
3. Упорядочивание элементов образовательного процесса на формальной базе коммуникативного аппарата языка (исследование составных частей образования, а также логических связей и отношений между ними) [13].

Получено представление относительно общей возможности применения функционального подхода к словообразованию в рамках педагогики. В то же время неясна внутренняя необходимость такого применения, поскольку оно осуществлено в результате формального соотнесения, а не на основании тождества смысловой структуры. Внутренне необходимые связи функционального подхода и педагогического исследования позволяет установить философская рефлексия, опирающаяся на принцип позитивной парадоксальности.

Принцип позитивной парадоксальности допускает особую интерпретацию философского знания. Это знание следует представлять в контексте интенции на предельность утверждения и потенции самоотрицания такой интенции в свете актуальной мультивариативности частых «пределов» (вера/разум, природа/человек и др.). Основной характеристикой философского знания выступает неустранимая противоречивость. Об этом свидетельствует история философии, в рамках которой явно прослеживается направленность на категоричность («общезначимость», предельность) суждения и невозможность действительной категоричности в силу собственной историчности философии. Поэтому в содержание понятия философского смысла некоторого явления действительности входит признак обязательности характеризовать данное явление с точки зрения предельности и историчности его бытия одновременно [19].

Такая позиция предполагает, что функциональный подход к словообразованию не просто способствует организации теории и практики педагогического процесса или служит целям его формального упорядочивания. Функциональный подход позволяет выделить особую иерархию уровней. Функции языка, характеризующие отдельные уровни, обусловливаются установками, которые свойственны существующим традициям (например, достаточно наивно говорить о домовых в современной повседневной жизни, а вот о «магических ядрах» в составе физики элементарных частиц вполне допустимо [20, с. 66-68]). Язык формируется в контексте усвоенных понятий, а также разработанных на данном основании способов терминологического отражения действительности. Такие способы позволяют обобщить педагогический процесс в инвариантных структурах и гарантируют внутреннюю его упорядоченность, обеспечивая обоснованность и возможность системного воспроизведения в рамках коллективной практики.

В то же время допущение философской составляющей предполагает в рамках функционального подхода критическое отношение к наличному знанию, определяя протекание педагогического процесса в границах оппозиционных по собственной сути отношений. Причем уровень оригинальности этапов процесса, сила их воздействия соответствует степени их оппозиционности. С математической точки зрения это можно представить в виде логарифмической функции, где на месте аргумента – количественные показатели образования новых понятий и способов их связи. Такие показатели определяют постепенный рост или, напротив, убывание инновационности, отражая в языке координацию традиций и инноваций и обусловливая усиление (ослабление) методологической культуры.

Таким образом, мы приходим к выводу о том, что трансформация смысла философии позволяет выявить возможности ее технологизации. Ведущую роль в данном процессе играет философия науки, которая создает условия для такой трансформации. Вместе с тем становится оправданной обращение к отраслевой комбинаторике направлений научного познания и частичный отказ от понимания философии как синтетического знания, в котором главной задачей является разрешение, а не порождение парадоксов.